\documentclass[journal,twoside,web]{ieeecolor}
\usepackage{generic}
\usepackage{cite}
\usepackage{amsmath,amssymb,amsfonts}
\usepackage{graphicx}
\usepackage{textcomp}

\usepackage{algorithmicx}               
\usepackage{algorithm} 
\usepackage{algpseudocode}  

\usepackage{array}
\usepackage{booktabs}
\usepackage{colortbl}
\definecolor{dark-gray}{gray}{0.7} 
\usepackage{multirow}

\def\BibTeX{{\rm B\kern-.05em{\sc i\kern-.025em b}\kern-.08em
    T\kern-.1667em\lower.7ex\hbox{E}\kern-.125emX}}
\begin{document}

\title{Faster Mean-shift: GPU-accelerated clustering for cosine embedding-based cell segmentation and tracking}

\author{Mengyang Zhao, Aadarsh Jha, Quan Liu, Bryan A. Millis, Anita Mahadevan-Jansen, Le Lu, Bennett A. Landman, Matthew J.Tyskac and Yuankai Huo* 
\thanks{M. Zhao is with Thayer School of Engineering, Dartmouth College, Hanover, NH 03755, USA}
\thanks{A. Jha, Q. Liu, B.A. Landman, and Y. Huo are with the Department of Electrical Engineering and Computer Science, Vanderbilt University, Nashville, TN 37235 USA}
\thanks{B.A. Millis and Matthew J.Tyskac are with the Department of Cell and Developmental Biology, Vanderbilt University, Nashville, TN 37235 USA}
\thanks{A. Mahadevan-Jansen is with the Department of Biomedical Engineering, Vanderbilt University, Nashville, TN 37235 USA}
\thanks{L. Lu is with the PAII Inc.,  Bethesda, MD 20817 USA}
}

\maketitle

\begin{abstract}
Recently, single-stage embedding based deep learning algorithms gain increasing attention in cell segmentation and tracking. Compared with the traditional "segment-then-associate" two-stage approach, a single-stage algorithm not only simultaneously achieves consistent instance cell segmentation and tracking but also gains superior performance when distinguishing ambiguous pixels on boundaries and overlaps. However, the deployment of an embedding based algorithm is restricted by slow inference speed (e.g., $\approx$1-2 mins per frame). In this study, we propose a novel Faster Mean-shift algorithm, which tackles the computational bottleneck of embedding based cell segmentation and tracking. Different from previous GPU-accelerated fast mean-shift algorithms, a new online seed optimization policy (OSOP) is introduced to adaptively determine the minimal number of seeds, accelerate computation, and save GPU memory. With both embedding simulation and empirical validation via the four cohorts from the ISBI cell tracking challenge, the proposed Faster Mean-shift algorithm achieved 7-10 times speedup compared to the state-of-the-art embedding based cell instance segmentation and tracking algorithm. Our Faster Mean-shift algorithm also achieved the highest computational speed compared to other GPU benchmarks with optimized memory consumption. The Faster Mean-shift is a plug-and-play model, which can be employed on other pixel embedding based clustering inference for medical image analysis. (Plug-and-play model is publicly available: {\color{red} https://github.com/masqm/Faster-Mean-Shift})
\end{abstract}

\begin{IEEEkeywords}
Mean-shift, GPU, Cell Tracking, Cell Segmentation, Embedding
\end{IEEEkeywords}

\section{Introduction}
\label{sec:introduction}
\IEEEPARstart {W}{ith} technical evolution in microscopy imaging, biomedical research has been advanced by spatial-temporal cell imaging, to understand cell motility and cell proliferation~\cite{b1}, embryonic development~\cite{b2}, tumorigenesis~\cite{b3} etc. To obtain dynamic information from acquired cell images and videos, manual tracing regarded as the gold standard of quantifying spatial-temporal microscope images. However, such a process is not only laborious and tedious but also impractical when terabyte (TB) level imaging data are acquired per day from a single imaging center~\cite{b42}. Therefore, automatic cell segmentation and tracking are crucial in cell image and video analyses, especially with the increasing spatial and temporal resolution from modern microscopy imaging.

Many computer-assisted segmenting and tracking methods ~\cite{b4,b5,b6} have been proposed over the past decades. Recently, with the explosive growth of artificial intelligence (AI) and deep learning technologies (e.g., convolutional neural network (CNN), and the long short-term memory model (LSTM)~\cite{b10,b11}), the performance of automatic cell segmentation and tracking has also been leveraged dramatically ~\cite{b7,b8,b9,b12,b13,b14}. Such methods were typically implemented as a "segment-then-track” two-stage paradigm, which linked segmented cells across frames via association (e.g., bipartite graph matching) algorithms. The advantages of such methods are (1) a clear problem definition, and (2) scalable to high resolution cell images.  However, the overall performance depends on both segmentation and association as independent tasks, with-out integrating the two synergetic tasks simultaneously. 

To aggregate the synergetic tasks as a holistic single stage algorithm, Payer et al.~\cite{b15} proposed a cosine embedding based recurrent stacked hourglass network (RSHN) for instance cell segmentation and tracking using microscope video sequences. They approached the instance segmentation and tracking problem from a new pixel-wise cosine embedding perspective to maximize the embedding similarity of the pixels within the same cell, while minimizing the embedding similarity across different cells. This method approached both the instance segmentation and tracking of cells within a single uniformed framework, which achieved superior performance compared with previous two-stage cell image processing approaches.

However, the major limitation of~\cite{b15}, which is a common issue in pixel embedding methods, is the slow inference speed when applying the trained model on testing images. For instance, the time for processing a single frame from the ISBI challenge data-set can take nearly two minutes per frame, which is a major bottleneck of deploying such algorithms on large-scale data. Based on our analyses, the majority of the computational time in ~\cite{b15} was spent on the clustering of embedding voxel-level features, as the time complexity of this algorithm is $O(T{n^2})$, where $T$ is the time for processing each pixel point and $n$ is the number of pixel points. Although there are several methods, such as adopting KD tree~\cite{b16} or ball-tree~\cite{b17}, could be employed to accelerate the algorithm. However, considering the large number of pixels in the image, the computational burden is much heavier than the 1D data in canonical clustering tasks, which leads to the considerably slow inference for medical image analysis. 

\begin{figure}[h]
\centerline{\includegraphics[width=\columnwidth]{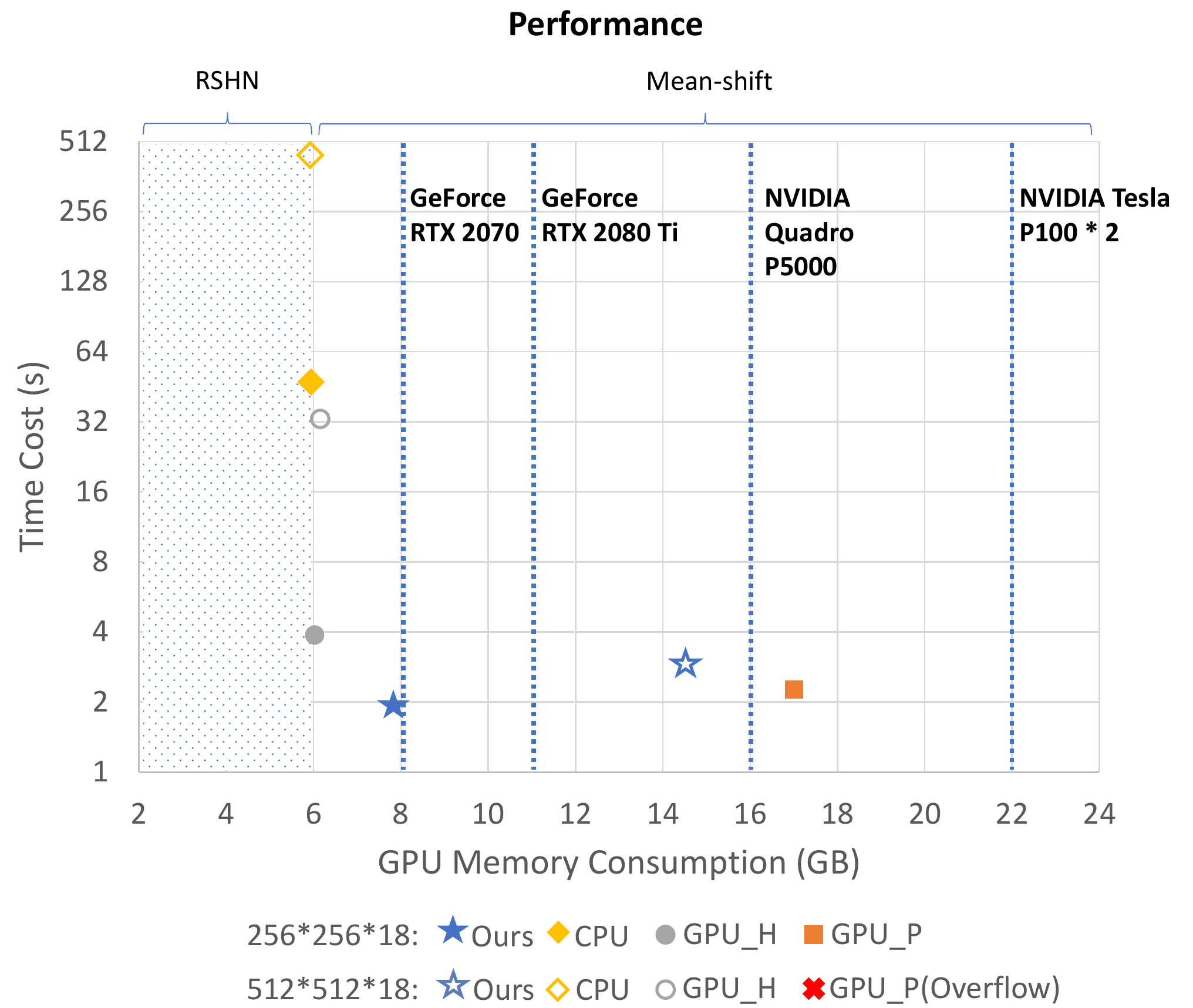}}
\caption{This figure shows the overall computational speed and GPU memory consumption when performing mean-shift clustering on simulated tensors with 256$\times$256$\times$18 and 512$\times$512$\times$18 resolution. The 256$\times$256 and 512$\times$512 indicates the image resolution for each frame, while 18 represents the dimensions of feature vectors of each embedded pixel. ``CPU'' represents the performance of ~\cite{b15}, which executed mean-shift only using CPU. ``GPU$\_$H''~\cite{b19} and ``GPU$\_$P''~\cite{b18} are two previously proposed GPU-accelerated fast mean-shift algorithms, where ``H'' and ``P'' indicate ``hybrid'' and ``parallel'' GPU accelerations respectively. The details are presented in Section 4.}
\label{fig1}
\end{figure}

Mean-shift is arguably the most widely used clustering algorithm in a large number of embedding based image processing, which is able to determine the number of clusters adaptively, as opposed to other clustering approaches (e.g., k-means~\cite{b29}) with a fixed number of clusters. In cell image processing, the mean-shift algorithm is proven to be more accurate (5-10\%,) than other clustering algorithms~\cite{b15,b28}. To accelerate the speed of mean-shift clustering, GPU accelerated algorithms with parallel computing have been proposed. For instance, the fast mean-shift algorithm ~\cite{b19} was developed to achieve significant speed-up compared with CPU based mean-shift clustering. Recently, ~\cite{b18} further accelerated computational speed with parallel tensor operations has been achieved. However, ~\cite{b18} is memory extensive, which is infeasible for processing high resolution image frames (e.g., 512$\times$512 in Figure \ref{fig1}) using the ordinary GPU cards. Therefore, a faster GPU accelerated clustering method with reasonable GPU memory consumption, is imperative for embedding based medical image analysis, 

Inspired by a parallel k-means algorithm~\cite{b37}, we believe that it is not necessary to compute feature vectors for all pixels when dealing with embedding based clustering. In this paper, we propose a novel GPU tensor accelerated mean-shift clustering algorithm, called Faster Mean-shift, to speed up the recurrent neural network (RNN) based cosine embedding framework for holistic cell instance segmentation and tracking. To optimize the GPU memory consumption, the online seed optimization policy (OSOP) and early stopping strategy are proposed to reduce unnecessary computing. The simulation as well as four real cohorts from the ISBI cell tracking challenge~\cite{b20} are employed in this study, to evaluate the accuracy, time cost, and GPU memory consumption of the proposed and baseline algorithms. We integrated the proposed Faster Mean-shift algorithm into the state-of-the-art RSHN framework ~\cite{b15}, to achieve 7-10 times speed-up during the inference stage, without sacrificing accuracy. The testing results show that our algorithms achieved the best computational speed with optimized memory consumption (Figure \ref{fig1}).

In summary, the main contributions of this study are as follows:

(1) We propose a novel Faster Mean-shift algorithm, which accelerates the embedding clustering based one-stage holistic cell instance segmentation and tracking.

(2) We propose the new online seed optimization policy (OSOP) and early stopping strategy to achieve the best computational speed with optimized memory consumption, compared with previous GPU-based benchmarks. ~\cite{b18,b19}.

(3) The proposed Faster Mean-shift achieved 7-10 times speed-up compared with the state-of-the-art embedding based cell instance segmentation and tracking algorithm ~\cite{b15}.

(4) Comprehensive simulations for in-depth theoretical analysis, as well as empirical validations on four ISBI cell tracking challenge data-set~\cite{b20} have been performed in this study.

The rest of the paper is organized as follows. In Section 2, we introduce background and related research relevant to cell segmentation and tracking. In Section 3, the mean-shift clustering and our proposed acceleration methods are presented. It includes the Faster Mean-shift algorithm and the theoretical derivation of OSOP. Section 4 focuses on presenting the implementation details and experimental results. Then, in Section 5 and 6, we provide ablation studies and conclude our work.

Our research is closely related to holistic instance segmentation and tracking ~\cite{b15}, as well as GPU accelerated clustering algorithms~\cite{b18,b19}. We present a brief introduction to the related research in this section.

\subsection{Instance Image Segmentation and Tracking}
Fully automatic image instance segmentation and tracking plays a critical role in biomedical image analyses, such as quantifying cells, cell nuclei, and other sub-millimeter structures from microscope images. Currently, the de facto standard instance segmentation and tracking methods are based on a deep learning algorithms ~\cite{b20}.

Deep learning can be traced back to 1998 when LeNet-5~\cite{b10}, a basic CNN model, was proposed by LeCun et al., where the convolutional layers were proposed to extract more generalizable image features than traditional feature engineering ~\cite{b25}. Then, a series of improvements, such as the ReLU activation function in AlexNet~\cite{b22} and the region selection algorithm in R-CNN~\cite{b23}, greatly improved the network recognition performance for processing the real-world image data. Different from processing instant image data, recurrent neural networks such as the LSTM~\cite{b11} and gated recurrent units (GRU)~\cite{b24} are proposed to incorporate temporal information into the deep learning framework.

Among different deep learning algorithms, the embedding based cell instance segmentation and tracking approach~\cite{b15} is the most related ones to our study. Meanwhile, without doing tracking, some other works focused on embedding based instance cell segmentation~\cite{b40,b41}. For embedding based methods, one major computational bottleneck is to cluster pixel-wise feature vectors to generate the final instances.

\begin{figure}[h]
\centerline{\includegraphics[width=\columnwidth]{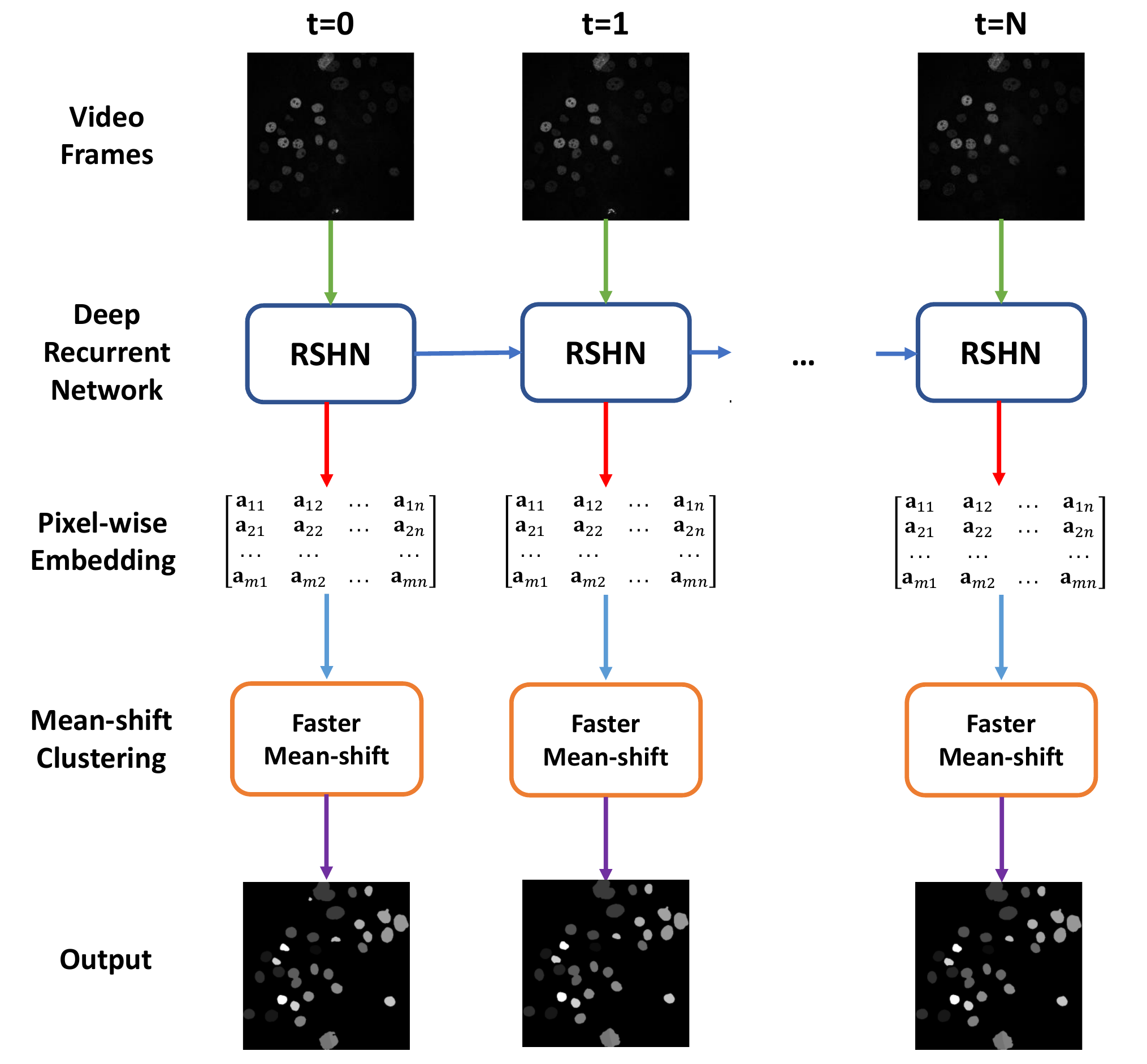}}
\caption{This figure presents the overall framework of a pixel embedding based holistic instance segmentation and tracking algorithm, with the proposed plug-and-play Faster Mean-shift GPU accelerated clustering algorithm.}
\label{fig2}
\end{figure}

\subsection{Single-stage Solution with Cosine Embedding}
Recently, instance cell segmentation and tracking have received increasing attention in many public challenges, such as the ISBI cell tracking challenge~\cite{b20}. Traditionally, instance segmentation and object tracking were performed separately as a two-stage design. In 2019, ~\cite{b15} integrated the instance segmentation and cell tracking into a holistic stacked hourglass network~\cite{b27} with pixel embedding based design. The well-known mean-shift clustering algorithm~\cite{b21} was employed during the inference stage to achieve final cell instance segmentation and tracking results.

The RSHN model~\cite{b15} was developed from the convolutional GRUs (ConvGRUs)~\cite{b26}, and the stacked hourglass network~\cite{b27}. It used a ConvGRU with 3$\times$3 filters and 64 outputs between the temporal paths to represent states and stacked two hourglasses in a row to improve network predictions. Based on the RSHN framework, each pixel in a video frame was converted to a high-dimensional embedding vector. The embedding vectors from different cell instances were distinguished by cosine similarity~\cite{b36}. For example, $\bf{A}$ was the embedding vector of pixel a, and $\bf{B}$ was the embedding vector of pixel b. The cosine similarity of $\bf{A}$ and $\bf{B}$ is defined as:

\begin{equation}
cos\left( {{\bf{A}},{\bf{B}}} \right){\rm{  = }}\frac{{{\bf{A}} \cdot {\bf{B}}}}{{\left\| {\bf{A}} \right\|{\rm{ }}\left\| {\bf{B}} \right\|}}
\label{eq1}\end{equation} which ranged from -1 to 1, where 1 indicates that two vectors have the same direction, 0 indicates orthogonal, and -1 indicates the opposite. Then, if $cos\left( {{\bf{A}},{\bf{B}}} \right)$ was $\approx1$, pixel a and pixel b were likely from the same instance. $\approx0$ was likely from different instances. After converting a video frame to an embedding vector matrix, a clustering algorithm was used to classify the pixels into different clusters/instances. According to ~\cite{b15}, the results obtained by the mean-shift algorithm were better than other clustering algorithms~\cite{b28}. The tracking and segmentation accuracy was improved 5-10\%, which demonstrated the advantages of the mean-shift algorithm.

\subsection{Fast Mean-shift Algorithms}
Mean-shift is an iterative clustering algorithm, which determines the number of clusters adaptively, as opposed to other clustering approaches (e.g., k-means~\cite{b29}). However, the major limitation of the mean-shift algorithm is the intensive computation, with a time complexity of $O(T{n^2})$. $T$ is the number of iterations for processing each data point and $n$ is the total number of data points in the data-set~\cite{b21}. Thus, due to the large number of pixels in image processing tasks, utilizing mean-shift clustering requires a large amount of computational time.

Several mean-shift acceleration methods have been developed to accelerate the mean-shift algorithm. These methods are generally divided into two families: CPU acceleration and GPU acceleration. For CPU acceleration, one major improvement is to reduce the number of vectors required in the mean-shift vector calculation. Traditionally, the calculation of the mean-shift requires all vectors from all the pixels. Therefore, the computational speed is accelerated if the algorithm is performed on a sub-sample of the feature vectors. To achieve better vector management, the fast searching algorithm proposed by Chalela et al.~\cite{b32} introduced a grid based searching paradigm. Another work~\cite{b16} used the KD tree to manage the vectors and further accelerate the searching approach. Based on the KD tree~\cite{b16} and ball-tree~\cite{b17}, sklearn~\cite{b33} provided a CPU accelerated mean-shift clustering implementation.

Although the aforementioned searching algorithms greatly improve speed, the overall processing speed of the CPU version is still not satisfactory for high dimensional features. For instance, the time for processing a 256$\times$256 pixels image could take more than one minute, and even longer for a higher resolution image. Therefore, GPU accelerated solutions were proposed to further speed up the mean-shift. Huang et al.~\cite{b19} proposed a hybrid CPU/GPU acceleration for mean-shift clustering. In their algorithm, GPU was used for calculating the mean-shift vector. However, because of the high time complexity of the region fusion step in their algorithm, the speedup for large image clustering yields inferior performance, at only 1.9 times. Huang et al~\cite{b18} proposed a fully parallel mean-shift algorithm. Instead of only using a GPU for vector calculation, in their algorithm, all vectors were shifted at the same time. This complete parallelization greatly sped up the algorithm. In the best case, their mean-shift clustering was approximately 40 times faster than the CPU version. However, this algorithm is limited by intensive resource consumption, such as GPU memory and CUDA cores, to complete parallel computing. Therefore, in their paper, large GPU clusters(with 64 worker GPUs and 384 GB GPU memory) were required to achieve the best acceleration. Thus, the consumption of computing resources could be unbearable for normal single GPU scenarios.

Indeed, in our implementation of  ~\cite{b18}, the memory resources of a single GPU card (12 GB GPU Memory) are quickly exhausted for the video embedding with 18 features and 512$\times$512 image resolution. Therefore, it is necessary to develop a new acceleration method, which has memory consumption that is acceptable for normal GPU cards, with even faster computational speed. Inspired by a parallel k-means algorithm~\cite{b37}, we realized that it is not necessary to calculate all vectors when dealing with clustering problems. Therefore, unlike Huang et al.~\cite{b18}, who parallelized all the vectors, we propose a Faster Mean-shift clustering algorithm based on adaptively determining a subset of vectors for computing.

\section{Proposed Method}
Our proposed Faster Mean-shift method is presented in this section. First, we introduce the mean-shift clustering for cosine embedding. Then, we present the proposed Faster Mean-shift algorithm with detailed theories and implementations.

\subsection{Mean-shift Clustering for Cosine Embedding}
The mean-shift algorithm is one of the most popular vector-based clustering methods, which is unsupervised and training-free~\cite{b29}. Assume there is a given vector ${{\bf{x}}_g}$ in a vector set $S = \{ {{\bf{x}}_1},{{\bf{x}}_2},...,{{\bf{x}}_n}\} $ of unlabeled data. The standard form of the estimated kernel density function $\hat f(x)$ at ${{\bf{x}}_g}$ is given by the following formula:

\begin{equation}
\hat f({\bf{x}}) = \frac{1}{{n{h^d}}}\sum\limits_{i = 1}^n k \left( {\frac{{d({{\bf{x}}},{{\bf{x}}_i})}}{h}} \right)
\label{eq2}\end{equation}where $k({\bf{x}})$ is a kernel function, $d({\bf{x}})$ refers to the distance function, and $h$ is referred to as the kernel bandwidth.

A standard kernel function is the Epanechnikov kernel~\cite{b30} given by the following formula:

\begin{equation}
k({\bf{x}}) = \left\{ {\begin{array}{*{20}{c}}
1&{\left\| {\bf{x}} \right\| \le 1}\\
0&{\left\| {\bf{x}} \right\| > 1}
\end{array}} \right.
\label{eq3}\end{equation} 

The Euclidean distance function between two vectors is:

\begin{equation}
d({{\bf{x}}_1},{{\bf{x}}_2}) = \left\| {{{\bf{x}}_1} - {{\bf{x}}_2}} \right\|
\label{eq4}\end{equation}

In the mean-shift clustering algorithm, the mean-shift vector is derived by calculating the gradient of the density function.

\begin{equation}\label{eq5}
\begin{split}
\nabla \hat f({\bf{x}}) &= \frac{1}{{n{h^d}}}\sum\limits_{i = 1}^n k \left( {\frac{{d({{\bf{x}}},{{\bf{x}}_i})}}{h}} \right) \\
&= \frac{1}{{n{h^d}}}\sum\limits_{i = 1}^n {\nabla k} \left( {\frac{{d({{\bf{x}}},{{\bf{x}}_i})}}{h}} \right) \\
&= \frac{2}{{n{h^{d + 2}}}}\left[ {\sum\limits_{i = 1}^n k \left( {\frac{{d({\bf{x}},{{\bf{x}}_i})}}{h}} \right)} \right] \\ 
&\times \left[ {\frac{{\sum\limits_{i = 1}^n {{{\bf{x}}_i}k} \left( {{{\left( {\frac{{d({\bf{x}},{{\bf{x}}_i})}}{h}} \right)}^2}} \right)}}{{\sum\limits_{i = 1}^n k \left( {{{\left( {\frac{{d({\bf{x}},{{\bf{x}}_i})}}{h}} \right)}^2}} \right)}} - {\bf{x}}} \right] \\
\end{split}
\end{equation}

Then, the content within the ``$[\cdot]$'' in the Eq.\eqref{eq5} is the mean-shift vector, presented as the following expression:

\begin{equation}
{M_h}({\bf{x}}) = \frac{{\sum\limits_{i = 1}^n {{{\bf{x}}_i}k} \left( {{{\left( {\frac{{d({\bf{x}},{{\bf{x}}_i})}}{h}} \right)}^2}} \right)}}{{\sum\limits_{i = 1}^n k \left( {{{\left( {\frac{{d({\bf{x}},{{\bf{x}}_i})}}{h}} \right)}^2}} \right)}} - {\bf{x}}
\label{eq6}\end{equation}

For a subset of feature vectors satisfying $S_h = \{ {{\bf{x}}_i}|d({{\bf{x}}_g},{{\bf{x}}_i}) \le h,{{\bf{x}}_i} \in S\} $, we reform the mean-shift vector as:

\begin{equation}
{M_h}({\bf{x}}) = {m_h}({\bf{x}}) - {\bf{x}}
\label{eq7}\end{equation} where the sample mean ${m_h}({\bf{x}})$ is defined as:

\begin{equation}
{m_h}({\bf{x}}) = \frac{1}{m}\sum\limits_{{{\bf{x}}_i} \in {S_h}} {{{\bf{x}}_i}} 
\label{eq8}\end{equation}

The iterative processing of calculating the sample mean converges the data to modes, which are the predicted clustering patterns. The proof of its mathematical convergence is provided in ~\cite{b21,b31}. The iterative process of mean-shift clustering is depicted as Algorithm \ref{alg:alg1}

\begin{algorithm}[h]  
  \caption{Mean-shift Clustering}  
  \label{alg:alg1}  
  \begin{algorithmic}[1] 
    \Require 
      $h$: Bandwidth;  
      $S$: Vector set;  
    \Ensure  
      $modes$: The modes of each cluster;   
    
    \For{${\bf{x}} \in S$}
    \State \# Initialization for each vector
    \State ${{\bf{x}}_g} \gets {\bf{x}}$
    \State Create a window: Bandwidth: $h$, Center: ${{\bf{x}}_g}$
    
    \State \# Mean-shift iteration
        \While{${{\bf{x}}_g}$ not converge}
            \State ${{\bf{x}}_g} \gets {m_h}({{\bf{x}}_g})$
            \State Update window to the new center
        \EndWhile
        \State $modes$ append ${{\bf{x}}_g}$
    \EndFor
    \State Prune $modes$
  \end{algorithmic}  
\end{algorithm} 

It is worth mentioning that, in this paper, we mainly use the cosine distance function in mean-shift, where the equation \eqref{eq4} needs to be replaced with the cosine distance between two vectors:

\begin{equation}
d({{\bf{x}}_1},{{\bf{x}}_2}) = 1 - \left( {\frac{{{{\bf{x}}_1} \cdot {{\bf{x}}_2}}}{{\left\| {{{\bf{x}}_1}} \right\|{\rm{ }}\left\| {{{\bf{x}}_2}} \right\|}}} \right)
\label{eq9}\end{equation}

By using mean-shift clustering, the output cosine embedding vectors from the RSHN~\cite{b15} model are clustered into different instances. According to the experiments in paper~\cite{b15}, the combination of mean-shift clustering and RSHN provided accurate instance segmentation and tracking results. However, the efficiency of such a method is limited by the intensive time consumption when performing standard mean-shift clustering (see Section 4). 

\subsection{Faster Mean-shift Algorithm}
\subsubsection{GPU-based parallelization}
Inspired by mean-shift GPU parallelization~\cite{b18} and k-means parallel implementation~\cite{b37}, we propose Faster Mean-shift algorithm, a novel GPU accelerated parallel mean-shift algorithm. The core idea of our proposed algorithm is to adaptively determine the number of seeds with an early stopping strategy to reduce the number of iterations in the mean-shift computation. The pseudo-code of the algorithm is shown in Algorithm \ref{alg:alg2}.

\begin{algorithm}[h]  
  \caption{Faster Mean-shift Clustering}  
  \label{alg:alg2}  
  \begin{algorithmic}[1] 
    \Require 
      $h$: Bandwidth;  
      $S$: Vector set;  
    \Ensure  
      ${\bf{x}}-modes$: A vector-modes list 
      
    \State \# Seed Selection
    \State Evenly random select seed vector set $S_{seed} \in S$
    
    \State \# Parallelization with GPU
    \For{${\bf{x}}_{seed} \in S_{seed}$}
    
        \While{${{\bf{x}}_{seed}}$ not converge}
            \State ${{\bf{x}}_{seed}} \gets m({\bf{x}}_{seed}) \cdot {k_h}({\bf{x}}_{seed},{{\bf{x}}_i})$
        \EndWhile
        \State $modes$ append ${{\bf{x}}_{seed}}$
    \EndFor
    \State Prune $modes$
    
    \For{${\bf{x}} \in S$}
        \State Cluster ${\bf{x}}$ by the distance to $modes$
    \EndFor
  \end{algorithmic}  
\end{algorithm} 

The algorithm first selects a batch of seed vectors from the input vector set $S$. According to our settings, in general, $N$ vectors are randomly selected from $S$ to form a subset $S_{seed}$. Then, these batched seeds are pushed into the GPU to perform parallel computation in lines 4-9 in Algorithm \ref{alg:alg2}. The mean-shift iterations are performed for each seed vector simultaneously. To save communication time on the GPU side, our algorithm does not search which points belong to the $S_h$ set. Instead, shown on line 6 in Algorithm \ref{alg:alg2}, our algorithm uses $m({\bf{x}})$ to calculate the mean-shift vector with all other points and then multiplies it with the kernel function, ${k_h}({\bf{x}},{{\bf{x}}_i})$, to obtain the mean-shift vector. The $m({\bf{x}})$ and ${k_h}({\bf{x}},{{\bf{x}}_i})$ functions are given by the following formula:

\begin{equation}
m({\bf{x}}) = \frac{1}{m}\sum\limits_{{{\bf{x}}_i} \in S} {d({\bf{x}},{{\bf{x}}_i})} 
\label{eq10}\end{equation}

\begin{equation}
{k_h}({\bf{x}},{{\bf{x}}_i}) = \left\{ {\begin{array}{*{20}{c}}
1&{d({\bf{x}},{{\bf{x}}_i}) \le h}\\
0&{d({\bf{x}},{{\bf{x}}_i}) > h}
\end{array}} \right.
\label{eq11}\end{equation}

Next, the position of ${{\bf{x}}_i}$ is updated according to the mean-shift vector. If the change of the distance between two iterations is less than a threshold (typically $h/1,000$), then the computation for this seed vector ${{\bf{x}}_{seed}}$ has converged. After parallelly manipulating the batch of seed vectors in the GPU, the modes are obtained in the vector set $S$. Next, such modes are further pruned and merged if their distance is small. In the end, all vectors are clustered according to their distance to each mode to obtain the final result.

Since only the seed vectors need to be manipulated in parallel, the GPU memory consumption is dramatically reduced, which enables parallel computing for mean-shift clustering on only one GPU card. However, to segment the real data-set, two critical tasks need to be tackled: (1) to determine and adjust the number of seed vectors, and (2) to reduce seed convergence time.

\subsubsection{Online Seed Optimization Policy (OSOP)}
The number of seed vectors plays an important role in the mean-shift algorithm. If the number of seeds is too low, the clustering results might not cover all modes. On the other hand, too many seeds lead to large GPU memory consumption ~\cite{b18}.

As opposed to traditional methods, wherein number of seeds are fixed, we proposed the OSOP approach to determine the minimum number of seeds adaptively, based on the number of instances and the foreground area in each prediction. For cell instance segmentation, we hypothesize that (1) the sizes and spatial distributions of the cells are homogeneous, and (2) the area ratio between the foreground ${A_{foreground}}$ and entire image ${A_{image}}$ is $r$. Based on such a hypothesis, the distribution of a seed follows a binomial distribution, where the percentage of each instance ${A_{foreground}}$  is presented as:

\begin{equation}
{A_{foreground}} = A_{{\rm{instance}}} \times I = A_{\rm{image}} \times r
\label{eq12}
\end{equation} where $I$ is the number of cell instances and $A_{{\rm{instance}}}$ is the average area of cells. Therefore, the probability that one random seed located within a particular instance cluster $P_{seed}$ is 

\begin{equation}
P_{seed} = \frac{{{A_{\rm{instance}}}}}{{{A_{{\rm{image}}}}}} =  \frac{r}{I}
\label{eq13}
\end{equation}

From the probabilistic model, the probability of a seed being located in an instance is a Binomial distribution. As a result, if the total number of seeds is $N$, the probability that each cluster has at least one seed is:

\begin{equation}
P_{seed/cluster} = {\left( {1 - {{\left( {1 - \frac{r}{{I}}} \right)}^N}} \right)^I}
\label{eq14}
\end{equation}


However, the hypothesis of an equal size of the instances in Eq. \eqref{eq13} does not hold in the real world. Therefore, the probability $P$ of each cluster having at least one seed is lower than the upper bound:

\begin{equation}
P \le {\left( {1 - {{\left( {1 - \frac{1}{{2I}}} \right)}^N}} \right)^I} 
\label{eq15}
\end{equation}

\begin{equation}\label{eq16}
\ln \left( {1 - {{\left( {1 - \frac{1}{{2I}}} \right)}^N}} \right) \ge \frac{{\ln (P)}}{I}
\end{equation}

\begin{equation}\label{eq17}
{\left( {1 - \frac{1}{{2I}}} \right)^N} \le 1 - {e^{\frac{{\ln (P)}}{I}}}
\end{equation}

\begin{equation}\label{eq18}
N\ln \left( {1 - \frac{1}{{2I}}} \right) \le \ln \left( {1 - {e^{\frac{{\ln (P)}}{I}}}} \right)
\end{equation}

Therefore, for a desired $P$, the $N$ should satisfy:

\begin{equation}
N \ge \frac{{\ln \left( {1 - {e^{\frac{{\ln (P)}}{I}}}} \right)}}{{\ln \left( {1 - \frac{r}{{I}}} \right)}} = {N_{\min }}
\label{eq19}
\end{equation}

In mean-shift clustering, each instance/cluster should have at least one seed to ensure all instances are successfully recognized. Therefore, if we expect the probability $P$ that the seed number would be sufficient for the $I$ instance above (e.g., $\ge 99\%$), we should have at least $N$ seeds above the low bound $N_{\min }$, where the $I$ and $r$ are calculated from the predicted segmentation. 

In our algorithmic implementation, since the areas of all cell instances are not identical, we typically need a larger number of seeds to cover all instances. A constant coefficient $\alpha \ge 1$ is introduced to enlarge the minimal numbers of seed in our algorithm.

\begin{equation}
N \ge \alpha \cdot {N_{\min }}
\label{eq20}
\end{equation}


\begin{figure}[h]
\centerline{\includegraphics[width=\columnwidth]{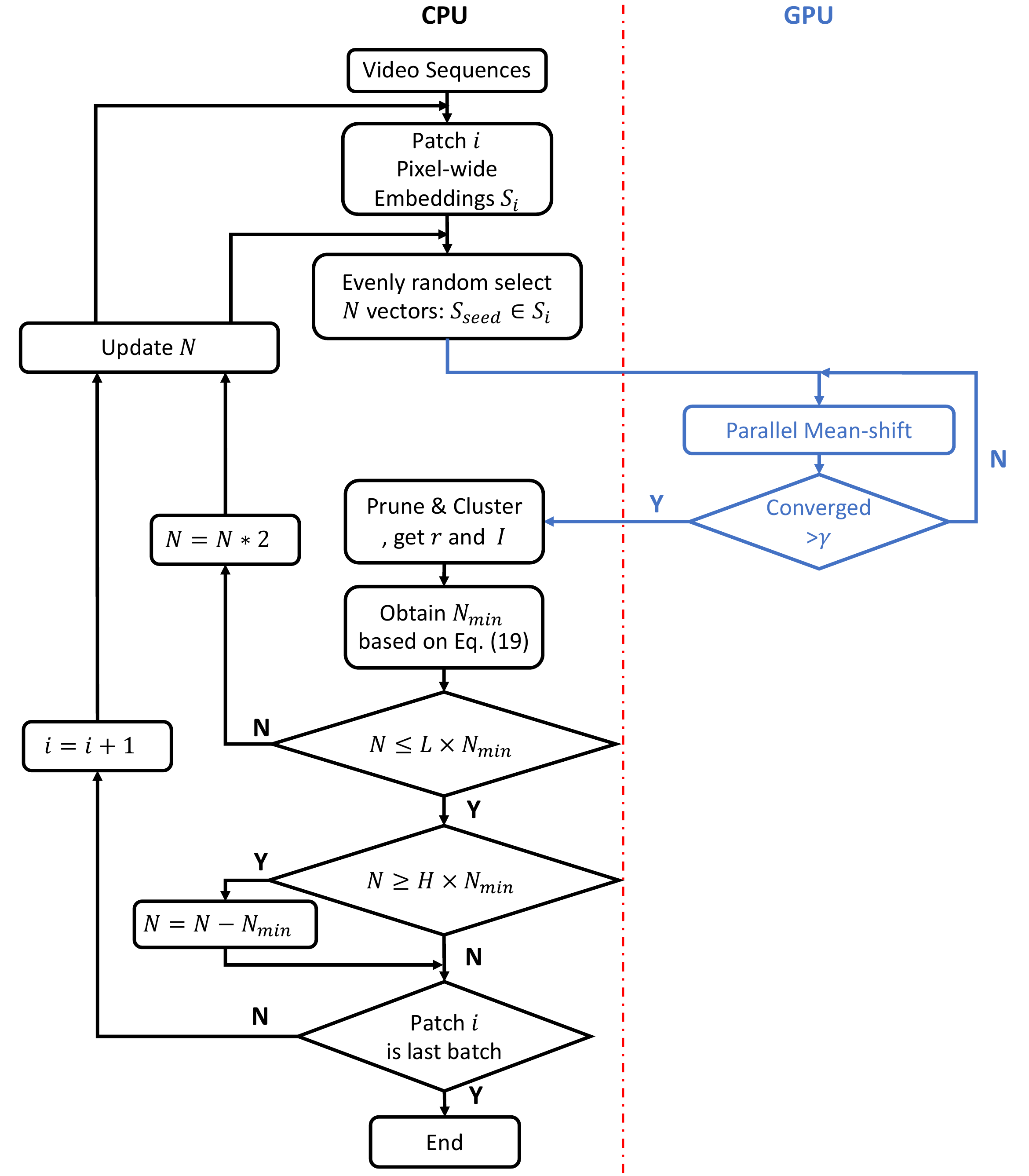}}
\caption{This figure depicts the flowchart of Faster Mean-shift. The black  portions of the diagram are processed by CPU, while the blue portions are processed by GPU.}
\label{fig3}
\end{figure}

\begin{figure*}[t]
\includegraphics[width=6.5 in]{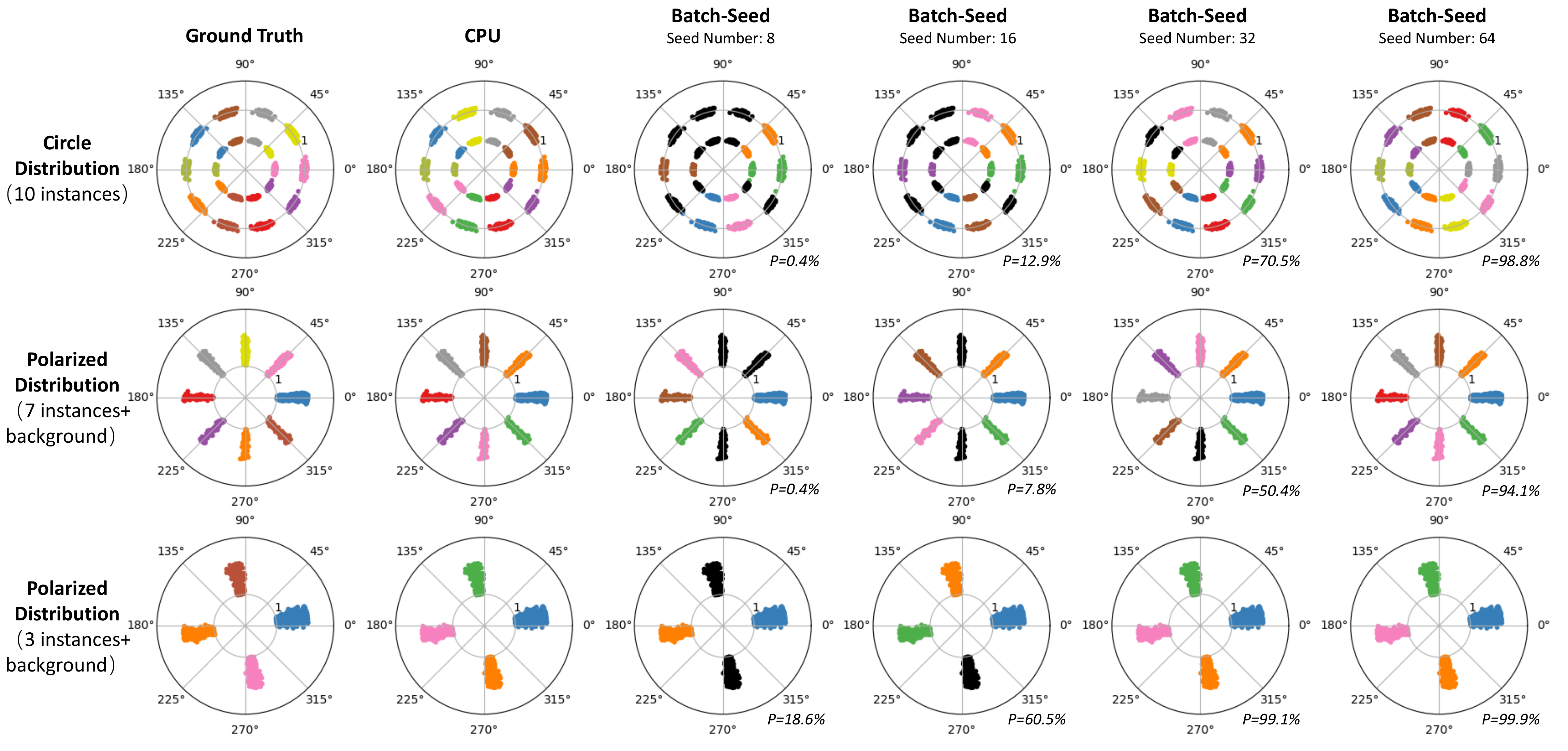}
\centering
\caption{This figure presents the simulation results using the proposed Faster Mean-shift algorithm. Each row indicates a distribution of the cosine embedding, while the columns represent the different number of seeds. In each subplot, different colors represent different clusters. Note that the blue points in the second and third rows simulate the image background, with more pixels. The black points indicate the failed clusters that are not captured by mean-shift. The theoretical probabilities of successful mean-shift clustering from OSOP are provided at the lower right corner of each subplot, which indicates the probability $p$ that each cluster contains at least one seed (Eq. \eqref{eq14}).}
\label{fig4}
\end{figure*}

\subsubsection{Early Stopping}
To further accelerate computational speed, we propose the early stopping strategy to optimize the total computational time for all the seed vectors. In an ideal situation, all seed vectors would converge simultaneously. However, the convergence is heterogeneous, where a few seeds will converge considerably slow and become the bottleneck of the entire GPU computing. To tackle this issue, we set a threshold percentage of converged seeds as $\gamma$. If more than $\gamma$ percentage of the seed vectors are converged, the mean-shift optimization of such iterations are terminated. The seed vectors that fail to convergence are discarded from the seed vector. In all experiments in this study, we empirically set the $\gamma=90\%$.

Consider Eq. \eqref{eq20}, the required minimal number of seeds for each iteration as:

\begin{equation}
N \ge \frac{\alpha }{\gamma } \cdot {N_{\min }} = L \cdot {N_{\min }}
\label{eq21}
\end{equation}

\noindent where $L \cdot {N_{\min }}$ is the minimal seed numbers for Faster Mean-shift implementation. 


\subsubsection{Faster Mean-shift Clustering.}
The flowchart of the entire Faster Mean-shift algorithm is shown in Figure \ref{fig3}. The computation is divided into two parts: the GPU and the CPU. The GPU portion mainly executes the iterative mean-shift computation in parallel (lines 4-9 in Algorithm \ref{alg:alg2}), while the CPU portion controls the number of seeds. 

The CPU is mainly responsible for adjusting the number of seed vectors. Initially, $N$ is set to 128 ($N_{initial}=128$). Then, during the following iterations, $N_{\min }$ will be updated based on the observed $I$ and $r$ in Eq. \eqref{eq19}. If $N$ is less than $L$ times of $N_{\min }$, the $N$ will be doubled for the next iteration. If $N$ is larger than $H$ times of the $N_{\min }$, the computational cost would be too expensive. In that case, $N$ is reduced as $N-N_{\min }$ for the next iteration. Based on our simulations, we empirically set $L=2$ and $H=8$ for all studies.

\section{Experiments and Results}
In this study, we performed both simulation and empirical validations via the ISBI cell tracking challenge data-set to evaluate the performance of the proposed Faster Mean-shift algorithm.

\subsection{Environments}
This research uses a standard\_NC6~\cite{b35} virtual machine platform at the Microsoft Azure cloud. The virtual machine includes one-half NVIDIA Tesla K80 accelerator~\cite{b34} card and six Intel Xeon E5-2690 v3 (Haswell) processors. The K80 GPU with 12 GB GPU memory was used in this study. The memory of the standard\_NC6~\cite{b35} virtual machine was 56 GB. The Faster Mean-shift algorithm was implemented with PyTorch and Python3. The source code of our proposed method has been publicly available $\footnote{https://github.com/masqm/Faster-Mean-Shift}$. To allow for a fair comparison on the same platform, we re-implemented the GPU accelerated mean-shift baselines~\cite{b18,b19} using PyTorch and Python3, which were originally implemented using the C language and OpenGL. The RSHN algorithm and CPU version of mean-shift were obtained from~\cite{b15}. During training, the learning rate was initially set to 0.0001, and decreases to 0.00001 after 10,000 iterations.


\subsection{Simulation}
The purpose of the simulation experiment is to evaluate the performance of the proposed Faster Mean-shift across different distributions and measure the costs of computational time and GPU memory. 

\subsubsection{Data}
Since the cosine similarity is used to distinguish vectors, we used polar coordinates to present the three distributions of simulated data as three rows in Figure \ref{fig4}.  In the circle distribution, 1,500 data points with embedding dimension = 2 were generated as two concentric circles with different radii. All data points were evenly distributed into 10 clusters. Gaussian noise with a standard deviation of 0.02 was added to the data. The second and third rows in Figure \ref{fig4} indicated polarized distributions with eight and four hidden clusters. Gaussian noise was added with a standard deviation of 0.005 (for the data with seven instances) and 0.01 (for the data with three instances). In the circle distribution (first row in Fig. \ref{fig4}), the intensities of Gaussian noise were evenly distributed from 0.95 to 1.05 (large circle) and 0.475 to 0.525 (small circle) on the polar-diameter directions, whose variations were determined by 10\% of the circles’ radius. In the simulations with three and seven instances (second and third rows in Fig. \ref{fig4}), the intensity is evenly distributed from 1 to 2. Different from even distribution in the first simulation, we simulated the proportion of the background in the real images, where the blue data-set had 750 points, accounting for 50\% of the total points. The remaining points were divided evenly across different clusters.


\subsubsection{Design}
We generated eight sets of simulation data with sizes of 1K, 2K, 5K, 10K, 20K, 50K, 100K, and 200K (1K=1,000), where each data point had 18 dimensions from ten randomly distributed clusters using $blobs$ function. For each set, the cluster number was set to 10, with a Gaussian noise that had a standard deviation of 0.01. The vectors were also orthogonal or opposite in different clusters to ensure distinct cosine similarities. We compared our algorithms with the CPU version of mean-shift (CPU)~\cite{b33}, hybrid CPU/GPU mean-shift (GPU$\_$H)~\cite{b19}, and fully parallel mean-shift (GPU$\_$P)~\cite{b18}. 

\begin{figure}[h]
\centerline{\includegraphics[width=\columnwidth]{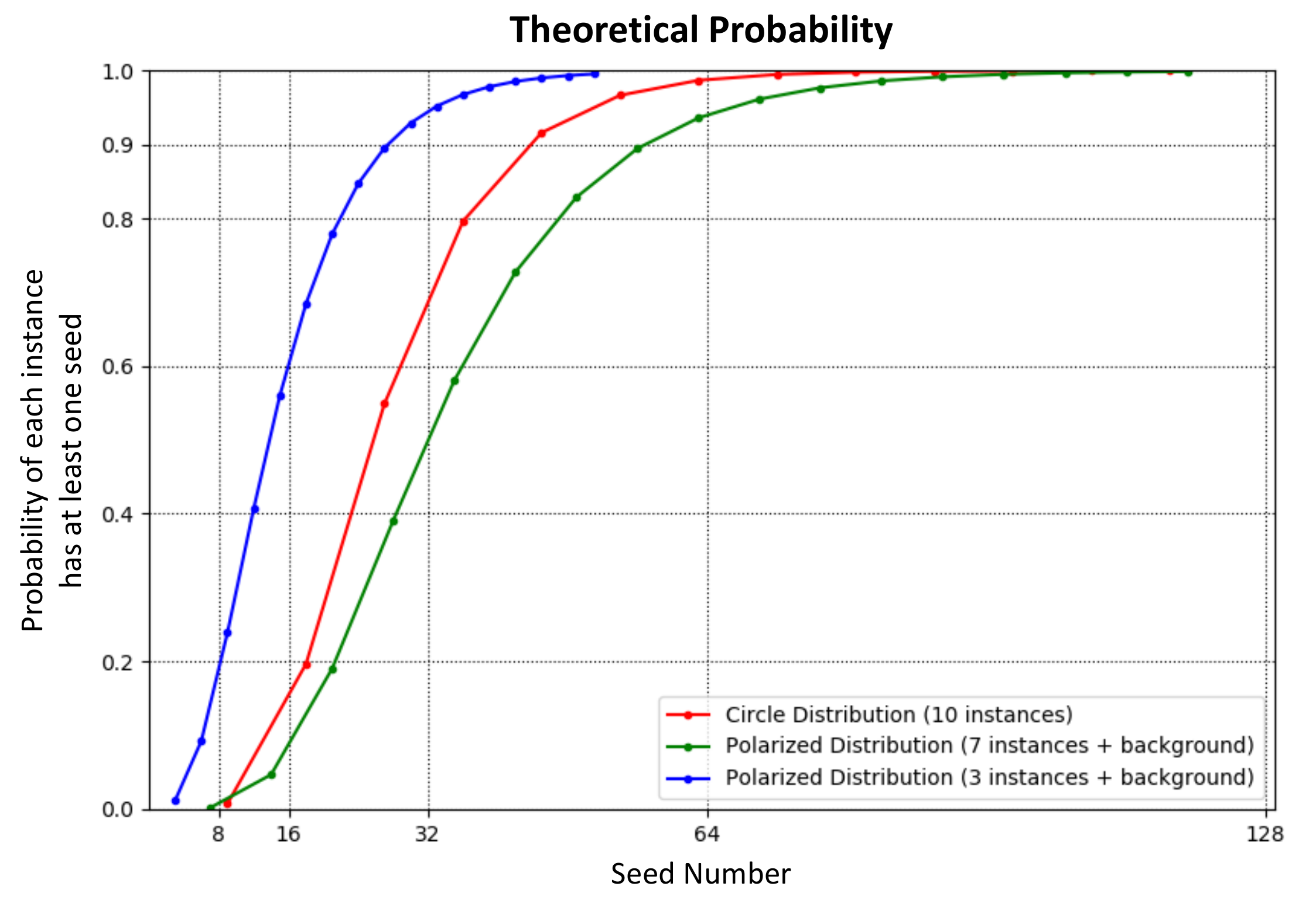}}
\caption{This figure presents the theoretical probability of correct clustering (each cluster has been assigned at least one seed) with different numbers of seeds for simulation. The three theoretical curves match the three simulation experiments in Figure. \ref{fig4}.}
\label{fig9}
\end{figure}

For the simulations (in Figure \ref{fig4}), the $h$ parameter was set to 0.1. We set different numbers of seeds for all experiments to evaluate the effects of such a hyper parameter. For the speed and memory test (in Table \ref{tab1}, Figure \ref{fig7} and \ref{fig8}), we recorded the computational time and GPU memory cost on each data-set using NVIDIA Management Library (NVML)~\cite{b38}. The peak GPU memory usage was reported as the GPU memory cost. To ensure robustness, we repeated the above experiment five times and reported the average measurements as the final results. 

\begin{table*}[t]
\caption{\textbf{Quantitative Results of Simulation}}
\setlength{\tabcolsep}{3pt}
\arrayrulecolor{black}
\renewcommand\arraystretch{1.3}
\centering
\begin{tabular}{p{3.3cm}<{\centering} | p{3cm}<{\centering} | p{1.2cm}<{\centering} p{1.2cm}<{\centering}  p{1.2cm}<{\centering} p{1.2cm}<{\centering} p{1.2cm}<{\centering}  p{1.2cm}<{\centering}  p{1.2cm}<{\centering} p{1.2cm}<{\centering}}
\hline
\hline
\multirow{2}{*}{\textbf{Method}} & \multirow{2}{*}{\textbf{Metric}} & \multicolumn{8}{c}{\textbf{Number of 2-D Vectors}}\\
\cline{3-10}
 & & 1K & 2K & 5K & 10K & 20K & 50K & 100K & 200K \\
\hline
\textbf{CPU} & Time Cost (s/frame) & \textbf{0.078} & \textbf{0.254} & \textbf{0.734} & 2.439 & 6.429 & 27.901 & 121.725 & 446.319 \\
~\cite{b33,b15} & GPU Memory (MB) & - & - & - & - & - & - & - & - \\
\hline
\textbf{GPU\_H}& Time Cost (s/frame) & 1.059 & 1.076 & 1.122 & \textbf{1.264} & \textbf{1.684} & 2.832 & 11.213 & 32.622 \\
~\cite{b19}  & GPU Memory (MB) & 216 & 216 & 216 & 216 & 216 & 218 & 238 & 236 \\
\hline
\textbf{GPU\_P} & Time Cost (s/frame) & 1.572 & 1.559 & 1.586 & 1.594 & 1.773 & 2.256 & 3.877 & 7.561 \\
~\cite{b18} & GPU Memory (MB) & 216 & 216 & 236 & 256 & 374 & 1174 & 3294 & 10156 \\
\hline
\multirow{2}{*}{\textbf{Ours}} & Time Cost (s/frame) & 1.624 & 1.618 & 1.664 & 1.698 & 1.747 & \textbf{1.892} & \textbf{2.107} & \textbf{2.692} \\
 & GPU Memory (MB) & 236 & 236 & 256 & 296 & 374 & 608 & 1004 & 1766 \\
\hline
\hline
\multicolumn{7}{l}{*The CPU version of mean-shift does not use the GPU, so GPU memory is not available(-).}\
\end{tabular}
\label{tab1}
\end{table*}

\subsubsection{Results}
\begin{figure}[h]
\centerline{\includegraphics[width=\columnwidth]{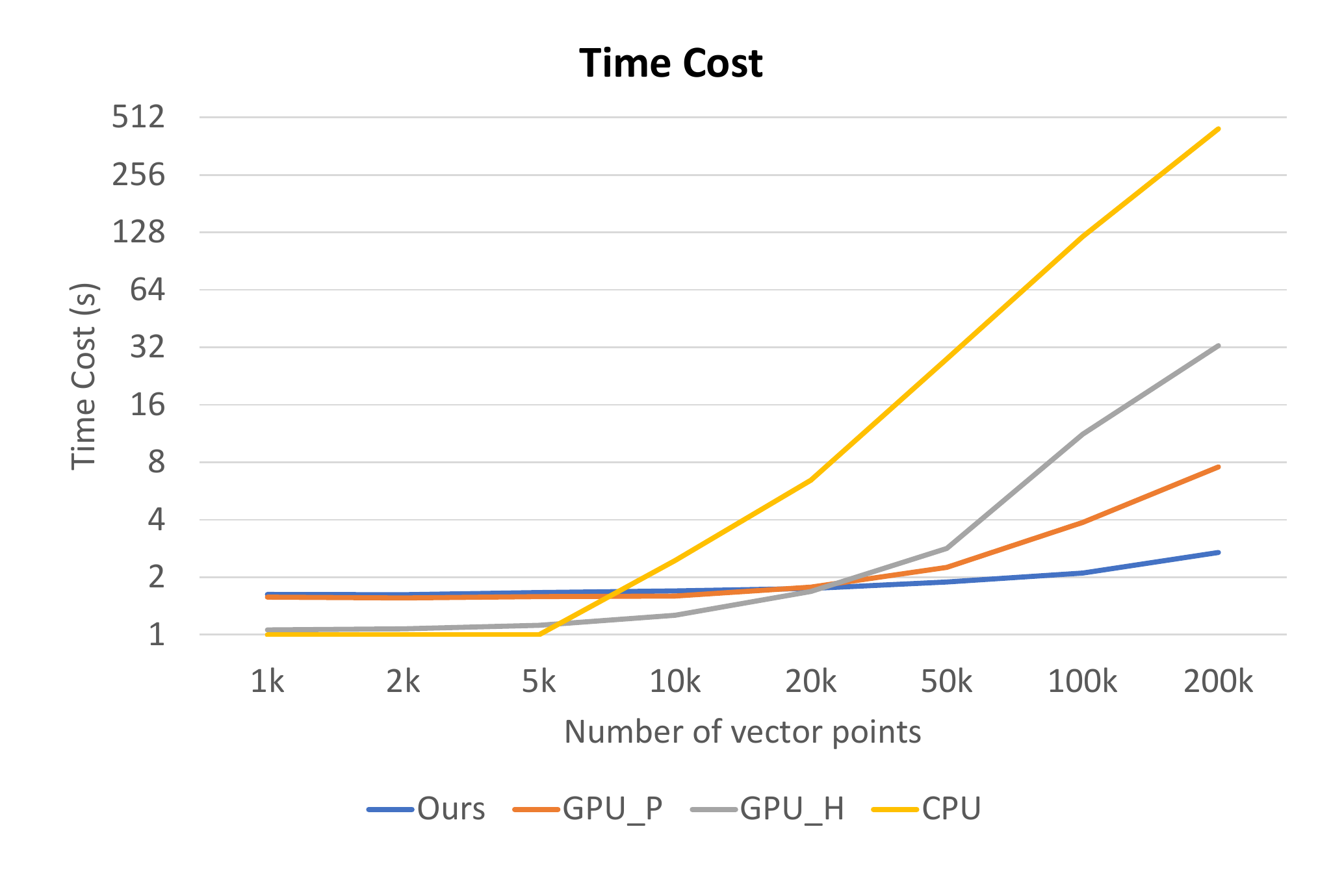}}
\caption{This figure shows costs of computational time for different methods using simulations.}
\label{fig7}
\end{figure}

\begin{figure}[h]
\centerline{\includegraphics[width=\columnwidth]{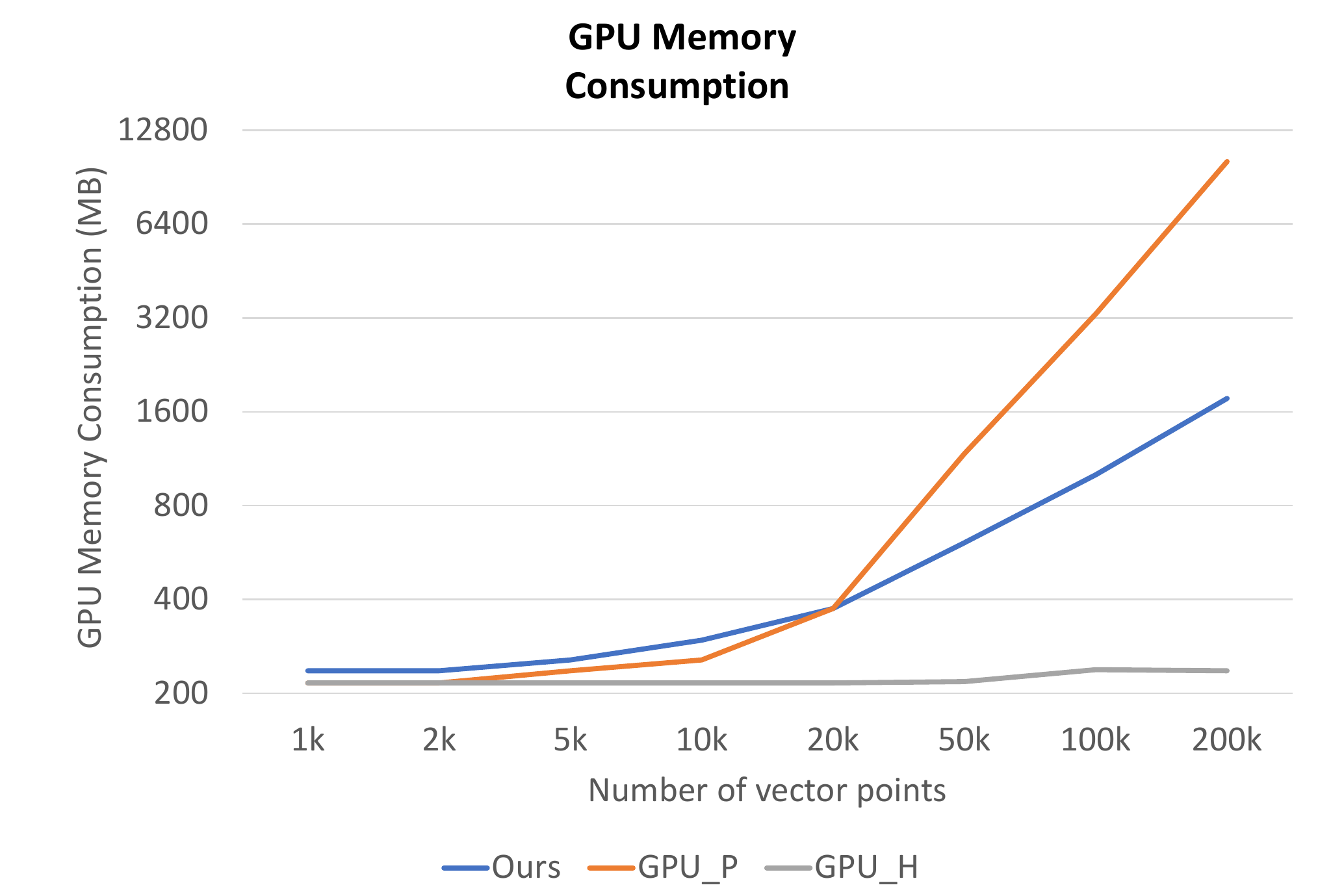}}
\caption{This figure shows GPU memory consumption of different methods using simulations. Since the CPU version does not use GPU, only three GPU related methods are presented.}
\label{fig8}
\end{figure}

The clustering results are presented in Figure \ref{fig4}, where the black color clusters indicate the failures. Figure \ref{fig9} shows the theoretical probability of successful instance segmentation across different seed numbers, which matches the simulation results in Figure \ref{fig4}. Meanwhile, we also evaluated the time cost (Table \ref{tab1} and Figure \ref{fig7}) and GPU memory cost (Figure \ref{fig8}) of the proposed Faster Mean-shift algorithm, compared with the baseline methods. In Figure \ref{fig4}), our algorithm correctly clustered the points in circle distributions when the number of seed is 64, which matched the theoretical curves in Figure \ref{fig9}. The remaining two simulations all matched their theoretical curves as well. The simulations demonstrated the number of seeds matched our theoretical derivations.


The costs of computational time and GPU memory are presented in Table \ref{tab1}, Figure \ref{fig7} and \ref{fig8}. Briefly, in Table \ref{tab1} and Figure \ref{fig7}, we tested our algorithm with different sizes of vectors and clusters, compared with the CPU~\cite{b33}, GPU$\_$ H~\cite{b19}, and GPU$\_$P~\cite{b18}. Figure \ref{fig7} shows that as the number of data points increased, the computational time of the CPU version increased dramatically. Within GPU algorithms, when the dataset was 10K-20K, the baseline methods performed slightly better than our algorithm. However, our algorithm achieved better speed performance after 20K. Considering a small 256$\times$256 image consists of more than 65K pixels, our algorithm has superior computational speed for image processing. 

GPU memory consumption results are presented in Figure \ref{fig8}. The GPU memory overhead of GPU$\_$H was the lowest. However, it yielded worse computational efficiency~\cite{b19}. For GPU$\_$P, as an opposite extreme case, which has large GPU consumption to achieve fast computational speed. Our Faster Mean-shift achieved the highest computational speed as well as had less had less memory consumption than GPU$\_$P. Therefore, for a standard single GPU card situation, the memory overhead of our algorithm was superior, especially for image processing. 

The simulations demonstrated the superior computational speed of our proposed method as well as the efficient GPU memory consumption.

\subsection{Empirical Validation}
In order to evaluate the performance of the Faster Mean-shift algorithm on the real-world cell image instance segmentation and tracking, we applied our method on four cohorts from the ISBI cell tracking challenge~\cite{b20}. 

\begin{figure*}[t]
\includegraphics[width=6.5 in]{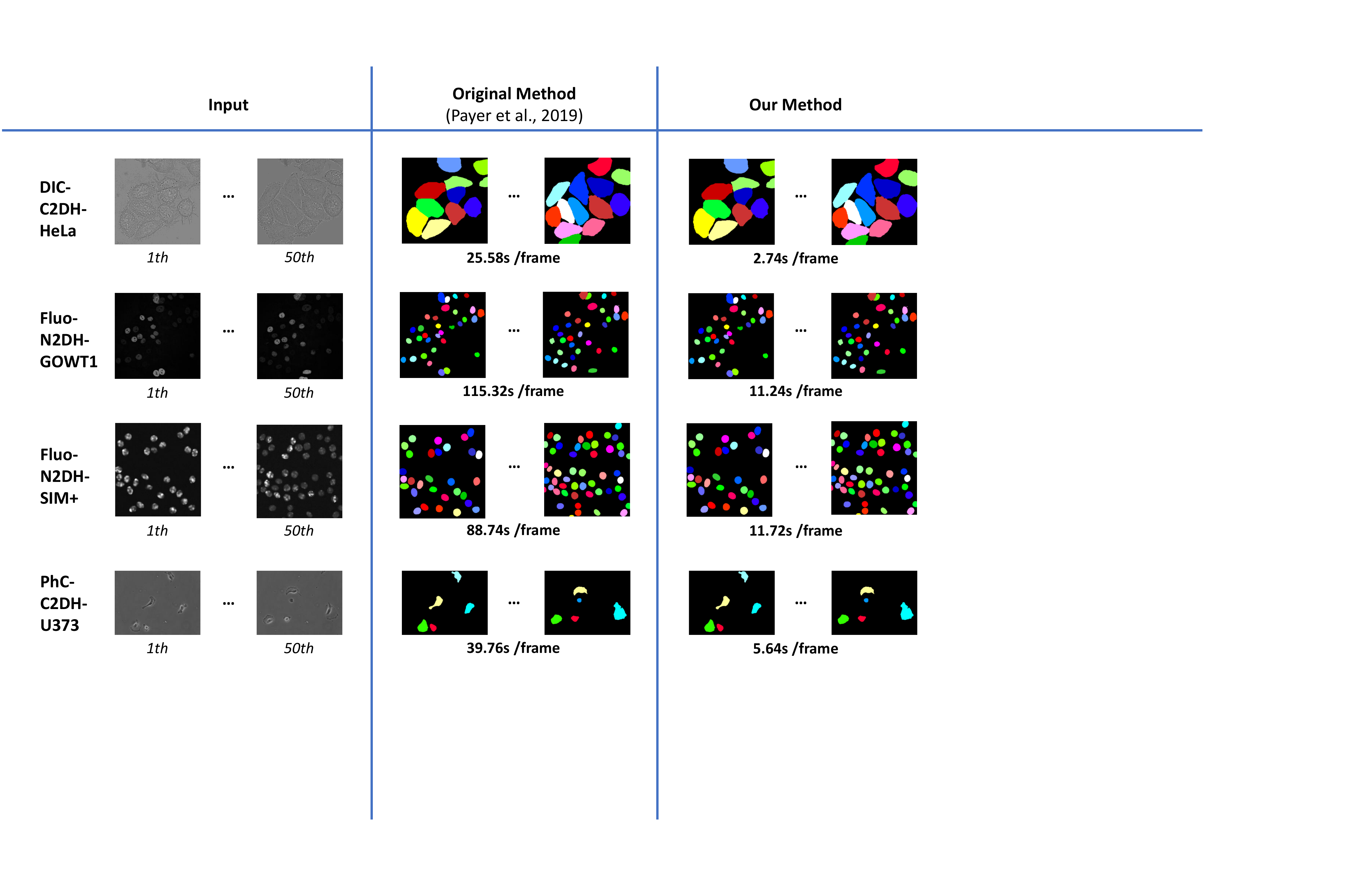}
\centering
\caption{This figure shows the qualitative instance segmentation and tracking results of the different methods. The left panel shows the input images. The middle panel shows the instance segmentation and tracking results of the baseline method. The right panel presents the results of the proposed method. The time costs per frame are also shown for different methods. }
\label{fig5}
\end{figure*}

\begin{table*}[!t]
\caption{\textbf{Quantitative Result of Empirical Validation}}
\setlength{\tabcolsep}{3pt}
\arrayrulecolor{black}
\renewcommand\arraystretch{1.3}
\centering
\begin{tabular}{p{3.5cm}<{\centering} | p{3.5cm}<{\centering} | p{2.4cm}<{\centering} p{2.4cm}<{\centering}  p{2.4cm}<{\centering} p{2.4cm}<{\centering} }
\hline
\hline
\multirow{2}{*}{\textbf{Method}} & \multirow{2}{*}{\textbf{Metric}} & \multicolumn{4}{c}{\textbf{Data-Set}}\\
\cline{3-6}
& & DIC-C2DH-HeLa & Fluo-N2DH-GOWT1 & Fluo-N2DH-SIM+ & PhC-C2DH-U373 \\
\hline
\textbf{Original}(CPU) & Time Cost (s/frame) & 25.58 & 115.32 & 88.74 & 39.76  \\
~\cite{b15,b33} & GPU Memory (GB) & 5.92 & 5.92 & 5.92 & 5.92  \\
\hline
\textbf{GPU\_H} & Time Cost (s/frame) &  9.83 &  37.54 &  18.60 & 10.81  \\
~\cite{b19} & GPU Memory (GB) & 6.16 & 6.16 & 6.16 & 6.16  \\
\hline
\textbf{GPU\_P} & Time Cost (s/frame) & - & - & - & -  \\
~\cite{b18} & GPU Memory (GB) &  overflow &  overflow &  overflow &  overflow  \\
\hline
\multirow{2}{*}{\textbf{Ours}} & Time Cost (s/frame) &  \textbf{2.74} & \textbf{11.24} & \textbf{11.72} & \textbf{5.64}  \\
 & GPU Memory (GB) &  8.38 & 7.24 &  6.95 & 6.82  \\
\hline
\hline
\multicolumn{5}{l}{*5.92GB is the minimal GPU memory consumption for the deep learning based feature extraction. }\\
\multicolumn{5}{l}{*Due to the GPU overflow, the time-cost is not available(-)}
\end{tabular}
\label{tab2}
\end{table*}

\begin{figure*}[t]
\includegraphics[width=6.5 in]{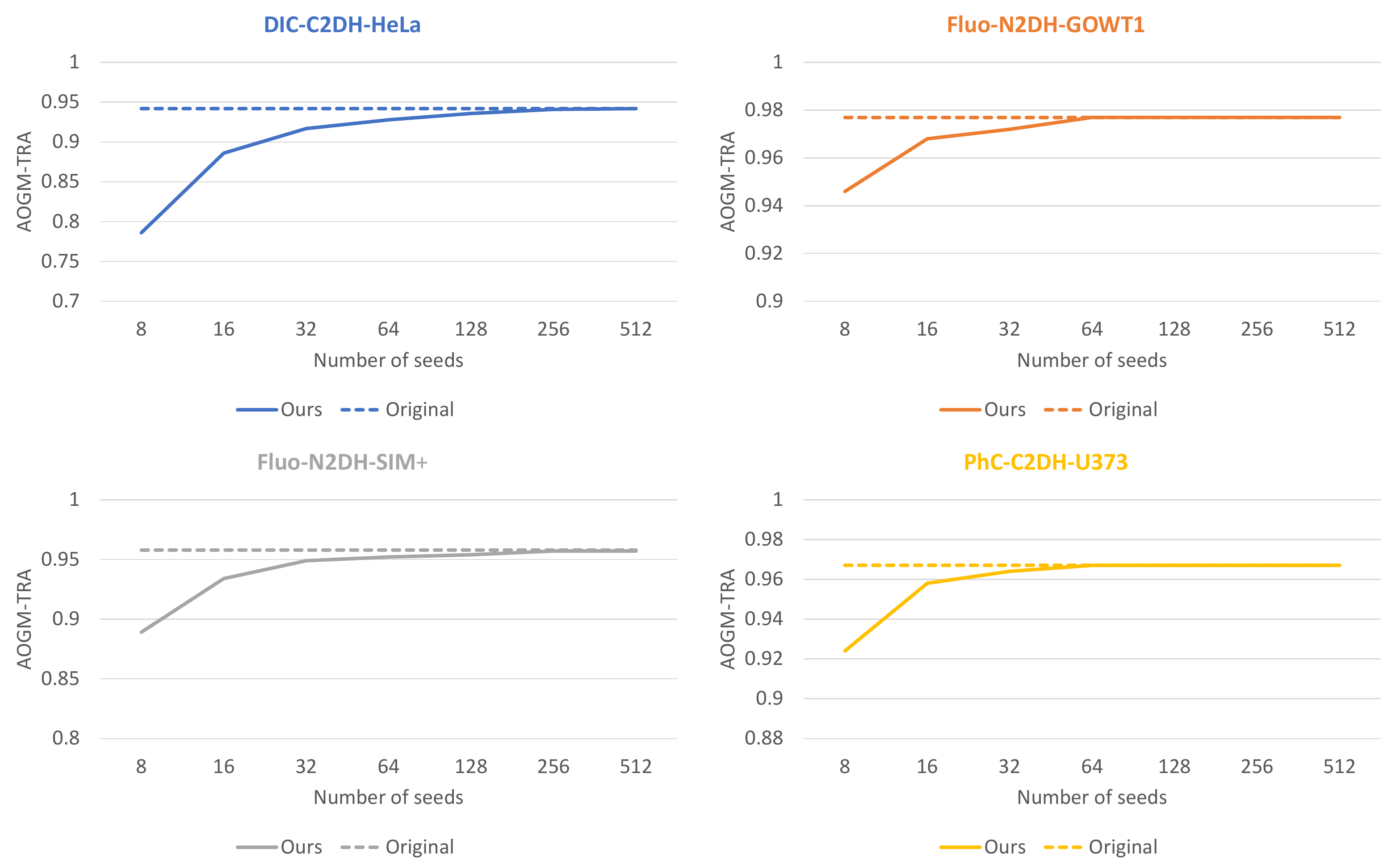}
\centering
\caption{This figure presents the standard TRA measurements ~\cite{b20} for the Faster Mean-shift (solid lines) and upper bound (dashed lines) when using all pixels for clustering.}
\label{fig6}
\end{figure*}

\subsubsection{Data}
In this testing, we used microscope video sequences from the ISBI cell tracking challenge~\cite{b20}. Four cell image datasets of different sizes, shapes, and textures were adopted to test the Faster Mean-shift clustering: (1) HeLa cells on a flat glass (DIC-C2DH-HeLa) with 512$\times$512 pixels, (2) GFP-GOWT1 mouse stem cells (Fluo-N2DH-GOWT1) with 1024$\times$1024 pixels, (3) Simulated nuclei of HL60 cells stained with Hoescht (Fluo-N2DH-SIM+) with 696$\times$520 pixels, and (4) Glioblastoma-astrocytoma U373 cells on a polyacrylamide substrate (PhC-C2DH-U373), with 660$\times$718 pixels. In addition, the testing results of remaining three data-sets are also shown in the Appendix B.

\subsubsection{Design}
The cell tracking and segmentation model proposed by Payer et al~\cite{b15}. with mean-shift clustering, was used as the benchmark. In the original model, the CPU version of mean-shift algorithm from sklearn~\cite{b33} was used to cluster the cosine embedding vectors generated by RSHN. We replaced the original CPU version mean-shift algorithm and used our proposed Faster Mean-shift algorithm for clustering. Moreover, we also replaced CPU version mean-shift with GPU$\_$H and GPU$\_$P (Ablation Studies). However, due to the limitation of GPU memory, when using GPU$\_$P, the GPU card encountered a memory overflow. As a result, speed performance and GPU memory consumption are not available. In the experiments, $h$ was set to 0.1 according to ~\cite{b15}. The $N_{initial}$ was set to 128 based on Figure \ref{fig9}. During the testing phase, 50 frames from each of the above four data-sets were employed for testing. We recorded the resource consumption by repeating the algorithm three times and reported the average results. 

\subsubsection{Results}
The qualitative results were shown in Figure \ref{fig5}, and the computational speed results were shown in Table \ref{tab2}. For the cell image data-sets, our algorithm achieved a 7-10 times speedup compared with \cite{b15}. 

In Figure \ref{fig6}, the normalized Acyclic Oriented Graph Matching measure for tracking(AOGM-TRA)\cite{b20},\cite{b39} was used as the accuracy metric. When the number of the seeds was higher than 256, the performance reached its upper bound limitation. The results demonstrated that the speed-up of our Faster Mean-shift did not sacrifice the accuracy.

\begin{table}[h]
\caption{\textbf{Time-cost (per frame) for Different $N_{initial}$}}
\setlength{\tabcolsep}{3pt}
\renewcommand\arraystretch{1.25}
\begin{tabular}{  p{3.25cm}<{\centering} p{1.5cm}<{\centering} p{1.5cm}<{\centering} p{1.5cm}<{\centering}}
\hline
\hline
$N_{initial}$ & =64 & =128 & =256\\
\hline
DIC-C2DH-HeLa & 2.59s & 2.74s & 2.98s\\
Fluo-N2DH-GOWT1 & 12.18s & 11.24s & 11.26s\\
Fluo-N2DH-SIM+ & 11.24s & 11.72s & 11.52s\\
PhC-C2DH-U373 & 5.66s & 5.64s & 5.57s\\
\hline
\hline
\end{tabular}
\label{tab3}
\end{table}

Moreover, the time cost for different $N_{initial}$ are provided in Table \ref{tab3}. $N_{initial}$ will affect the number of iterations in the first few frames. However, the influence of $N_{initial}$ on the average computational time is small as shown in Table \ref{tab3}. 

\begin{table}[h]
\caption{\textbf{Time-cost (per frame) for Different $H$}}
\setlength{\tabcolsep}{3pt}
\renewcommand\arraystretch{1.25}
\begin{tabular}{  p{3.25cm}<{\centering} p{1.5cm}<{\centering} p{1.5cm}<{\centering} p{1.5cm}<{\centering}}
\hline
\hline
$H$ & =4 & =8 & =16\\
\hline
DIC-C2DH-HeLa & 2.69s & 2.74s & 2.74s\\
Fluo-N2DH-GOWT1 & 11.02s & 11.24s & 11.30s\\
Fluo-N2DH-SIM+ & 11.36s & 11.72s & 11.81s\\
PhC-C2DH-U373 & 5.56s & 5.64s & 5.67s\\
\hline
\hline
\end{tabular}
\label{tab4}
\end{table}

$H$ will also affect the time-cost. In general, the required computational time per frame slightly increases with the larger $H$. And, the influence on average time-cost is shown in Table\ref{tab4}.

\section{Discussion}
\subsection{Performance Analyze}
The computational time cost per frame of different methods on different cohorts was shown in Table \ref{tab2}. For the cell image data-sets, our algorithm achieved the best computational efficiency across four cohorts. It is worth to mentioning that the speed improvement of total computational time in empirical validation is less than the simulation. The reason is that the time cost of an empirical validation consists of (1) the forward path of deep learning, and (2) the mean-shift clustering. However, in simulation, the time cost only contains the mean-shift clustering. As the forward deep learning path is included for the real data-set, the speed improvement of the total computational time in simulation is larger than the real data-set. However, considering that the RSHN required a considerable amount of time to generate cosine embedding, our algorithm achieved great improvement. 


The memory consumption using different methods on different cohorts was shown in Table \ref{tab2}. Our algorithm achieved much lower total GPU memory costs with GPU$\_$P, which was acceptable for a prevalent GPU card (e.g., GeForce RTX 2080 Ti with 11 GB GPU memory). Note that the 5.92GB GPU memory was required for all methods as this is the minimally required GPU memory for running RSHN feature extraction using TensorFlow.  

\subsection{Overlapping and Dense Cells}
As the cell tracking challenge data do not consist of a large number of heavily overlapped cells, we performed a simulation as shown in Figure \ref{figap1} in Appendix A to evaluate the performance of the proposed method at highly overlapped scenarios. From the results, the required theoretical $N_{\min }$ (from Eq. \eqref{eq19}) increased with higher overlapping ratios. The proposed method achieved accurate segmentation results using the same settings ($L=2$ and $H=8$) as other experiences across this study. The simulation shows that the proposed Faster Mean-shift algorithm is able to cluster objects with different overlapping ratios. 

Other issue is that the instance objects might not be uniformly distributed across the whole image, especially for high-resolution images. To tackle that challenge, the RSHN method has considered the non-uniform and dense distribution scenarios with patch-based design \cite{b15}, where the high-resolution images are split to multiple 256×256 image patches. Then the patches are processed and aggregated to the original resolution. The performance of the proposed method on dense object quantification is presented in PhC-C2DL-PSC dataset in Appendix C.

\subsection{Limitations}
There are several limitations of current cosine-embedding based instance cell segmentation and tracking as well as the proposed Faster Mean-shift. First, the higher image resolution images, especially in 3D imaging, might lead to significantly larger GPU memory consumption, and may not fit the current hardware. Second, one major limitation of the proposed method is that the method is designed for 2D cell imaging. However, the capability of processing 3D cell images would be critical, especially with the rapid development of 3D microscopy imaging. Therefore, a valuable future direction would be to extend the proposed method from 2D to 3D.  Third, even though the Faster Mean-shift accelerates the inference stage by a large margin, the speed is still not at a real time scale. Moreover, the cell tracking challenge data do not consist of a large number of heavily overlapped cells. However, it is important to compare the performance of the proposed method with highly overlapped objects versus minimal overlapping scenarios. Herein, we performed a simulation as shown in Fig. \ref{figap1} in Appendix A. 

\section{Conclusion}
In this study, we proposed a Faster Mean-shift algorithm for tackling the bottleneck of the state-of-the-art cosine embedding based cell instance segmentation and tracking. Compared with previous GPU-based mean-shift algorithms, our Faster Mean-shift method achieved better computational speeds, with acceptable memory consumption for a single ordinary GPU card. Using Faster Mean-shift, the processing speed for each frame was accelerated by 7-10 times compared to the state-of-the-art embedding based cell instance segmentation and tracking algorithm. As many recent studies have demonstrated the significant advantages and superior accuracy performance of embedding based methods, this algorithm provides a plug-and-play model, which is adapted for any pixel embedding based clustering inference. 

\section{Acknowledgement}
This study is supported by 
NSF Career Award 1452485 (Landman).

\bibliographystyle{IEEEtran}
\bibliography{references}{}


\onecolumn
\appendix

\section{Overlapping Simulation}
\begin{figure}[h]
\includegraphics[width=6.5 in]{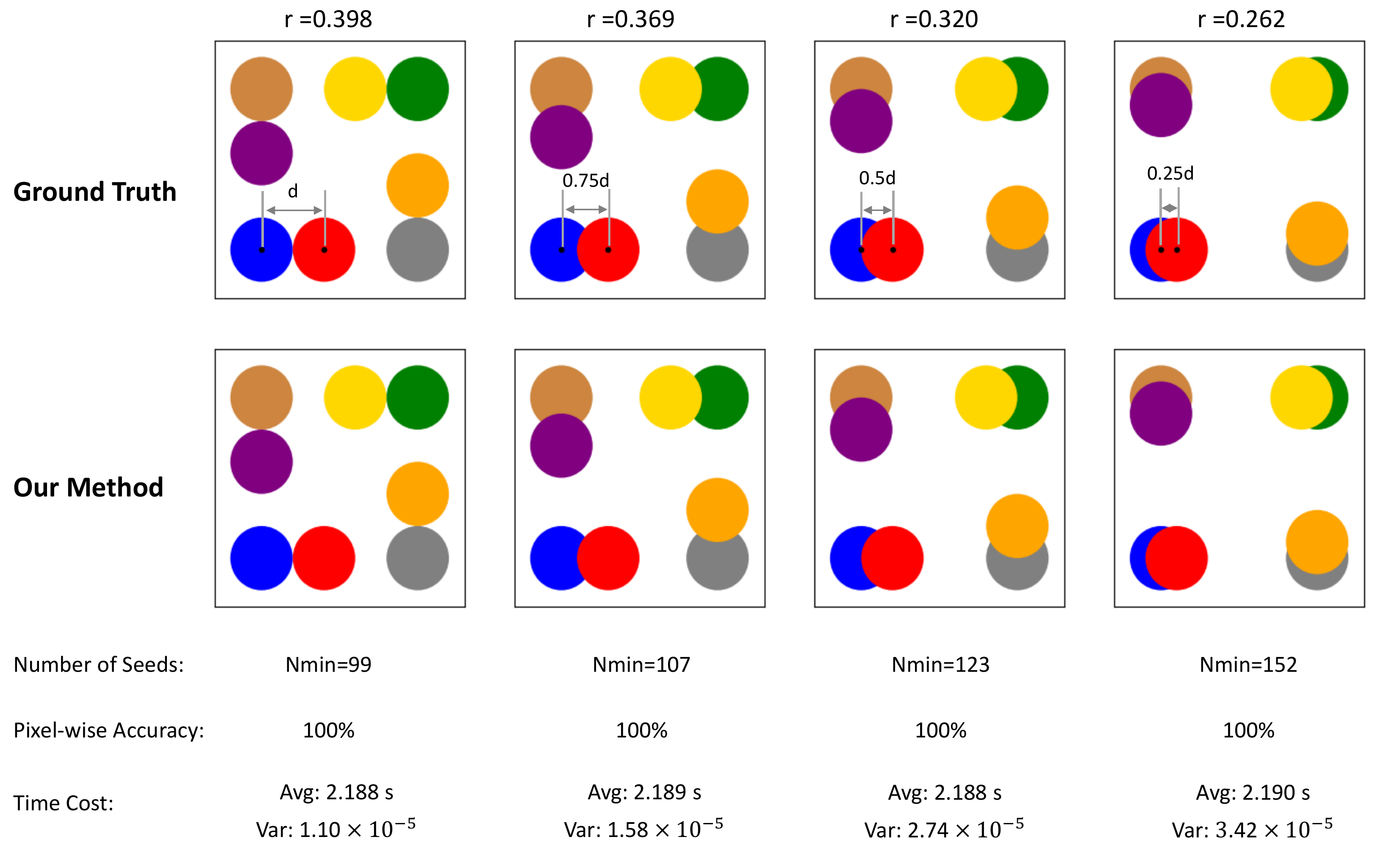}
\centering
\caption{This figure shows the simulation on different ratios of overlapping. The first row shows the ground truth of simulation. Different colors represent different cosine embeddings. The second row shows the results of the proposed method. The computational time of different simulations are also presented.}
\label{figap1}
\end{figure}

A simulation is provided to evaluate the theoretical numbers of seeds $N_{\min }$ based on the ratio between foreground and the entire image $r$ using \eqref{eq19} with probability P = 0.95. The resolution of the simulated data is 1000×1000×18, where 1000×1000 is the total numbers of feature vectors for an image and 18 is the dimension of cosine embedding for each vector. The simulated data are presented as color images for visualization in Fig.A.10, where each simulated data contains eight foreground circle objects (diameter d = 250 pixels) with different cosine embeddings. The different levels of overlapping are simulated by controlling the intersection between circles. The instance segmentation results using the theoretical $N_{\min }$ with different $r$ values are presented in Figure \ref{figap1}. From the results, the required $N_{\min }$ increased with a higher overlapping ratio. The computational times of different simulations are presented. The results showed that the default setting settings ($L=2$ and $H=8$) achieved the correct segmentation results. Note that this simulation did not include random noise, which is different from the ones in Figure \ref{fig4} Since this simulation is to evaluate the effects of overlapping, we avoid the potential entangled impacts from noise. 

\newpage
\section{Three Supplementary Data-sets}
\begin{figure}[h]
\includegraphics[width=6.5 in]{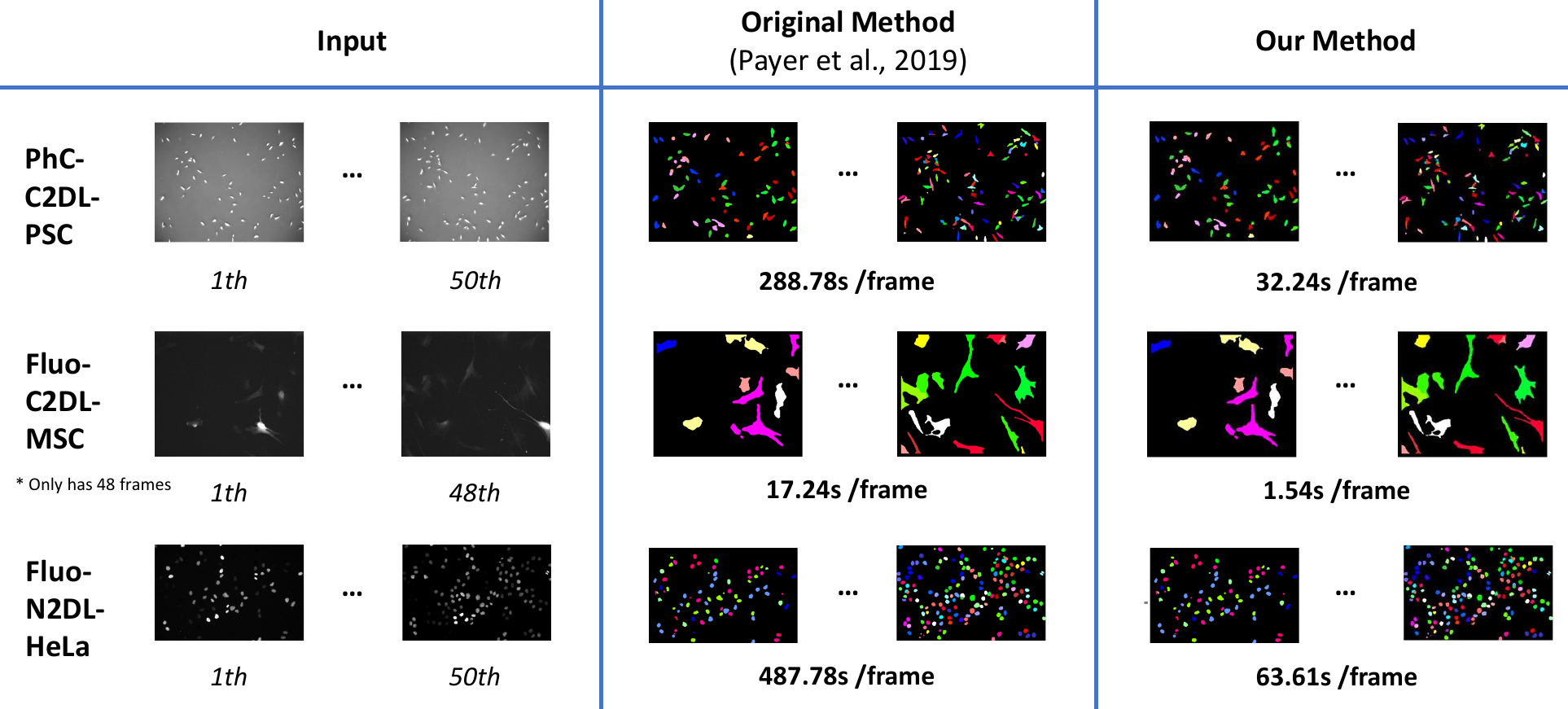}
\centering
\caption{This figure shows the qualitative instance segmentation and tracking results of three data-sets from Cell Tracking Challenge. The left panel shows the input video frames. The middle panel shows the instance segmentation and tracking results using the baseline method. The right panel presents the results using the proposed method. The time costs per frame are also shown for different methods. }
\label{figap2}
\end{figure}

\begin{table*}[h]
\caption{\textbf{Quantitative Result of Empirical Validation}}
\setlength{\tabcolsep}{3pt}
\arrayrulecolor{black}
\renewcommand\arraystretch{1.3}
\centering
\begin{tabular}{p{3.5cm}<{\centering} | p{3.5cm}<{\centering} | p{3cm}<{\centering} p{3cm}<{\centering}  p{3cm}<{\centering} }
\hline
\hline
\multirow{2}{*}{\textbf{Method}} & \multirow{2}{*}{\textbf{Metric}} & \multicolumn{3}{c}{\textbf{Data-Set}}\\
\cline{3-5}
& & PhC-C2DL-PSC & Fluo-C2DL-MSC & Fluo-N2DL-HeLa \\
\hline
\textbf{Original}(CPU) & Time Cost (s/frame) & 288.78 & 17.24 & 487.78 \\
~\cite{b15,b33} & GPU Memory (GB) & 5.92 & 5.92 & 5.92 \\
\hline
\textbf{GPU\_H} & Time Cost (s/frame) &  89.94 &  7.89 &  121.45 \\
~\cite{b19} & GPU Memory (GB) & 6.16 & 6.16 & 6.16  \\
\hline
\textbf{GPU\_P} & Time Cost (s/frame) & - & - & - \\
~\cite{b18} & GPU Memory (GB) &  overflow &  overflow &  overflow \\
\hline
\multirow{2}{*}{\textbf{Ours}} & Time Cost (s/frame) &  \textbf{32.24s} & \textbf{1.54} & \textbf{63.61} \\
 & GPU Memory (GB) &  7.49 & 6.21 &  8.58 \\
\hline
\hline
\multicolumn{5}{l}{*5.92GB is the minimal GPU memory consumption for the deep learning based feature extraction. }\\
\multicolumn{5}{l}{*Due to the GPU overflow, the time-cost is not available(-)}
\end{tabular}
\label{tabap1}
\end{table*}

\newpage
\section{Dense Cells Segmentation}
\begin{figure}[h]
\includegraphics[width=6.5 in]{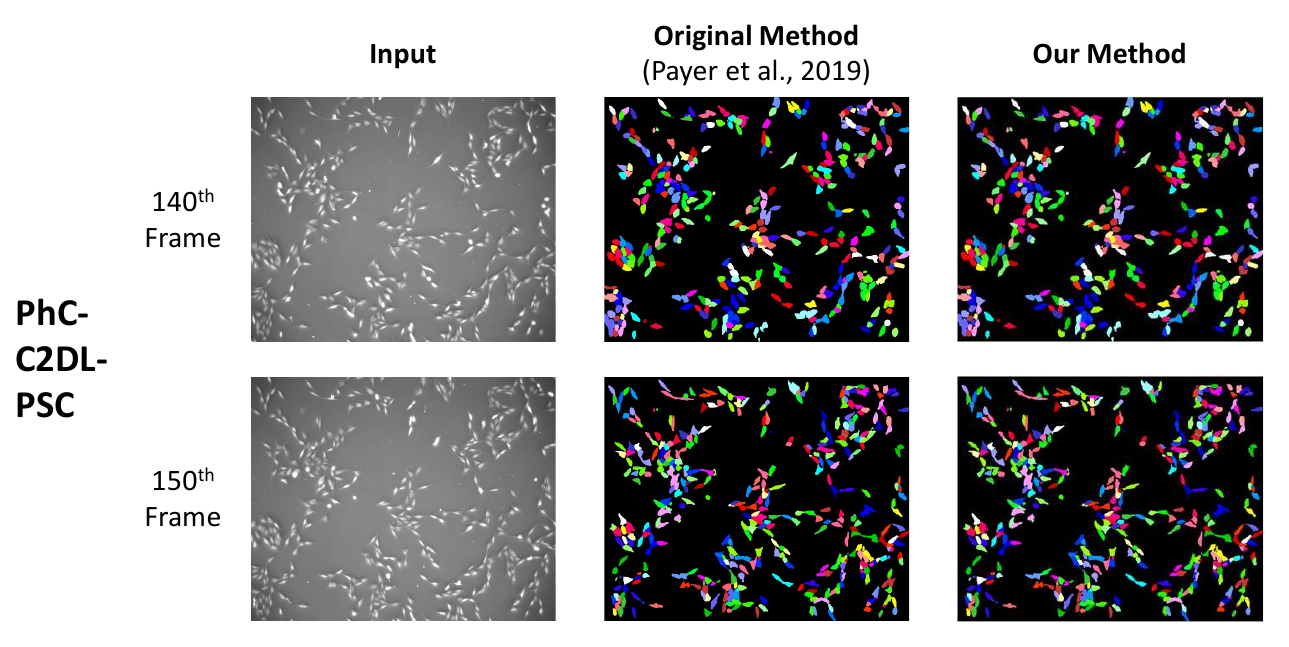}
\caption{This figure shows the qualitative results of dense object segmentation. }
\label{figap3}
\end{figure}

\end{document}